\documentclass[journal]{IEEEtran}

\usepackage{amsmath,amssymb,amsfonts}
\usepackage{soul}
\usepackage{pifont}
\usepackage{enumerate}
\usepackage{colortbl}
\usepackage{algorithm} 
\usepackage{algpseudocode}
\usepackage{graphicx}
\usepackage{textcomp}
\usepackage{tikz,caption}
\usepackage{pgfplots}
\pgfplotsset{compat=1.17}
\usepackage{xcolor}
\usepackage[inline]{enumitem}
\usepackage{float} 
\usepackage{cite}
\usepackage{url} 

\newlength{\TenSpaces}
\ifCLASSINFOpdf

\else

\fi

\begin{document}
\title{Comparison of Autoencoder Encodings for ECG Representation in Downstream Prediction Tasks}
\author{Christopher J. Harvey, Sumaiya Shomaji, Zijun Yao, \IEEEmembership{Member, IEEE}, Amit Noheria
\thanks{This work was supported by the University of Kansas Medical Center.}
\thanks{Christopher J. Harvey and Amit Noheria are with the Department of Cardiovascular Medicine, KU Medical Center, Kansas City, KS 66160 USA (e-mail: cjh@ku.edu; noheriaa@gmail.com)}
\thanks{Sumaiya Shomaji and Zijun Yao are with the Department of Electrical Engineering and Computer Science, University of Kansas, Lawrence, 
KS 66045 USA (e-mail: shomaji@ku.edu; zyao@ku.edu).}}

\maketitle

\begin{abstract}
The electrocardiogram (ECG) is an inexpensive and widely available tool for cardiovascular assessment.
Despite its standardized format and small file size, the high complexity and inter-individual variability of ECG signals (typically a 120,000-size vector at 1000hz) make it challenging to use in deep learning models, especially when only small datasets are available.
This study addresses these challenges by exploring feature generation methods from representative beat ECGs, focusing on Principal Component Analysis (PCA) and Autoencoders to reduce data complexity.
We introduce three novel Variational Autoencoder (VAE) variants—Stochastic Autoencoder (SAE), Annealed $\beta$-VAE (A$\beta$-VAE), and cyclical $\beta$-VAE (C$\beta$-VAE)—and compare their effectiveness in maintaining signal fidelity and enhancing downstream prediction tasks.
The A$\beta$-VAE achieved superior signal reconstruction, reducing the mean absolute error (MAE) to 15.7$\pm$3.2$\mu$V, which is at the level of signal noise.
Moreover, the SAE encodings, when combined with ECG summary features, improved the prediction of reduced Left Ventricular Ejection Fraction (LVEF), achieving an area under the receiver operating characteristic curve (AUROC) of 0.901.
This performance nearly matches the 0.910 AUROC of state-of-the-art CNN models but requires significantly less data and computational resources.
Our findings demonstrate that these VAE encodings are not only effective in simplifying ECG data but also provide a practical solution for applying deep learning in contexts with limited-scale labeled training data.
\end{abstract}


\section{Introduction}
\label{sec:introduction}

\IEEEPARstart{T}{he} electrocardiogram (ECG) is a non-invasive clinical tool that records the electrical activity of the heart through electrodes placed at standard locations on the skin.
In clinical practice, the time-varying electrical potential recorded between different standardized cathode and anode pairs is depicted as a 12-lead ECG.
ECG is a ubiquitous tool used for diagnosis of various health conditions.
Standard 12-lead ECG recording is done over 10 seconds at sampling rate of 500-1000 Hz thus producing 60,000 or 120,000 datapoints of signal data, spanning 8-17 cardiac cycles at normal heart rates.
Each normal cardiac cycle includes 3 sequential ECG waves: P wave (depolarization of the atria), QRS complex (depolarization of ventricles), and T wave (ventricular repolarization).
The characteristics of these waves—such as amplitude, frequency, and duration—convey information on cardiac structure and function, and varies heavily between different leads and among different individuals.

\textbf{Complexity of ECG Signal}: Raw ECG signal data is, especially in context of a limited number of training samples, difficult to use as an input for Deep Learning (DL) models to predict specific health diagnoses.
The intrarecording and interindividual variance in this data is large.
The P wave is short, low frequency and low amplitude; QRS complex is short, high frequency and high amplitude while the T wave is long, low frequency and intermediate amplitude.
These waves and intervals between cardiac cycles are interspersed with periods of zero electrical activity (or electrical baseline) which makes the signal data distribution very skewed.
All humans have unique hearts with variations in size, anatomy and electrophysiology, accounting for the interindividual differences in the ECG signal.
E.g., one person might have a QRS complex amplitude of 0.25 mV and another 5 mV.
The morphology of each person's ECG can be very different, e.g., an ECG lead of a person might have a smooth monophasic positive QRS complex (R wave) while another might have a notched R wave while another has triphasic Q-R-S deflections.
Age, sex, body structure, the specific lead in question and cardiac diseases all affect the morphology of the ECG.
The temporal volatility of occurrence of heartbeats during the 10-sec ECG recording, with additional possibility of abnormal cardiac rhythms, add further variability to the ECG signal.

\textbf{Challenge in Deep Learning}: 
The complexity of ECG data requires more complex learning models, and the more complex the model, the more data it requires to generalize.
The prediction targets of clinical significance, like reduced cardiac contractile function or Left Ventricular Ejection Fraction (LVEF), may occur in only a fraction of the population.
The rarity of the prediction targets requires large or specialized datasets for training.
This means to use any advanced DL algorithm to predict such health conditions using raw 10-sec ECG signal data, it requires a large amount of labeled training data for that outcome (\textgreater50,000 samples), or the model will overfit.
Considering that even large healthcare systems may only have a few thousand ECGs labelled with an important health event (e.g., myocardial infarction, pulmonary embolism, etc.), the use of raw ECG signal data as input to a DL model for these outcomes invariably leads to overfitting.
As a consequence, DL hasn't been successfully used at a wide scale for training classification tasks in limited-size ECG datasets.
Thus, there is a critical need to find a lower-dimensional representation of the complex raw ECG signal data to avoid overfitting and enable DL.

The historical method to simplify ECG into fewer variables is to use ECG summary statistics such as heart rate, P wave duration, PR interval, QRS duration, QT interval, QRS axis, and amplitudes of the P wave, QRS complex and T wave \cite{berkaya2018survey}.
However, the nuances of ECG morphology are not fully captured in these statistics.
For instance, two ECGs could have identical summary statistics but exhibit completely different morphologies, such as Left Bundle Branch Block (LBBB) versus Right Bundle Branch Block (RBBB).

In the literature, there have been comprehensive studies on reducing the dimensionality of ECG data for various purposes.
For example, Kumar and Chakrapani \cite{kumar2022classification} used PCA to simplify ECG signals before classification, and Dasan and Panneerselvam \cite{DASAN2021102225} employed a convolutional denoising autoencoder with LSTM for signal compression.
Wosiak \cite{Wosiak+2019+489+496} utilized PCA based on data characteristics for arrhythmia classification.
However, these methods suffer from the drawback of not fully capturing the non-linear relationships in ECG data, leading to potential loss of critical information necessary for accurate clinical diagnosis.

Variational Autoencoders (VAEs) offer advantages in addressing these limitations. VAEs are capable of:
\begin{itemize}
    \item \textbf{Non-Linear Feature Extraction}: VAEs can capture complex, non-linear relationships within the data due to their DL architecture.
    \item \textbf{Structured Latent Space}: The probabilistic nature of VAEs introduces regularization, leading to a continuous and smooth latent space facilitating generalization.
    \item \textbf{Generative Capabilities}: VAEs can generate new data samples, aiding in data augmentation, especially valuable when dealing with small training datasets.
\end{itemize}

Notably, van de Leur et al. \cite{10.1093/ehjdh/ztac038} utilized VAEs to improve the explainability of deep neural network-based ECG interpretation by learning underlying factors of variation in ECG morphology (the FactorECG).
Their approach aimed to enhance interpretability by providing a more transparent representation of ECG features.
However, while they used a VAE to encode the 12-lead representative-beat ECG, we experimented with Autoencoder (AE) and VAEs with different encoder-decoder architectures and loss functions to encode the 3-(X, Y, Z)-lead representative-beat ECG.
We also address the challenge of training DL models on limited data with high variability.

\textbf{Proposed Methodology}: To address the aforementioned challenges, specifically improving downstream predictions, we propose a VAE-based framework to reduce ECG data dimensionality.
Our approach differs from prior works by focusing on optimizing the latent space representations to both maintain high-fidelity signal reconstruction and enhance the performance of predictive models trained on small datasets.

We introduce three novel VAE variants—Stochastic Autoencoder (SAE), cyclical $\beta$-VAE (C$\beta$-VAE), and Annealed $\beta$-VAE (A$\beta$-VAE)—designed to address the specific challenges of encoding ECG data.
These models aim to improve the balance between reconstruction fidelity and latent space regularization, providing better representations for downstream tasks.

The following are the major contributions of this work:
\begin{itemize}
    \item A VAE-based framework for encoding time series data to reduce the complex ECG into a few latent encodings that retain the ECG signal's characteristics.
    \item Using VAE encodings as a practical method to apply DL in ECG datasets with limited training samples, which can be then used for traditional (e.g., tree-based) learning algorithms.
    \item Introduction of three novel VAE variants and a comprehensive comparison between them and standard methods.
\end{itemize}

\section{Methods}
\label{sec:formatting}

\subsection{Data}
We began by reducing the 10-sec ECG recordings to a single average representative beat, a 750 ms segment centered 100 ms after the onset of the QRS complex.
This reduction is useful because a 10-sec ECG typically contains multiple similar cardiac cycles (8 to 17 cycles at normal heart rates).
The representative beat captures the essential morphological features of the ECG while reducing the data size.

\begin{figure}[h]  
  \centering
  \includegraphics[width=1.\linewidth]{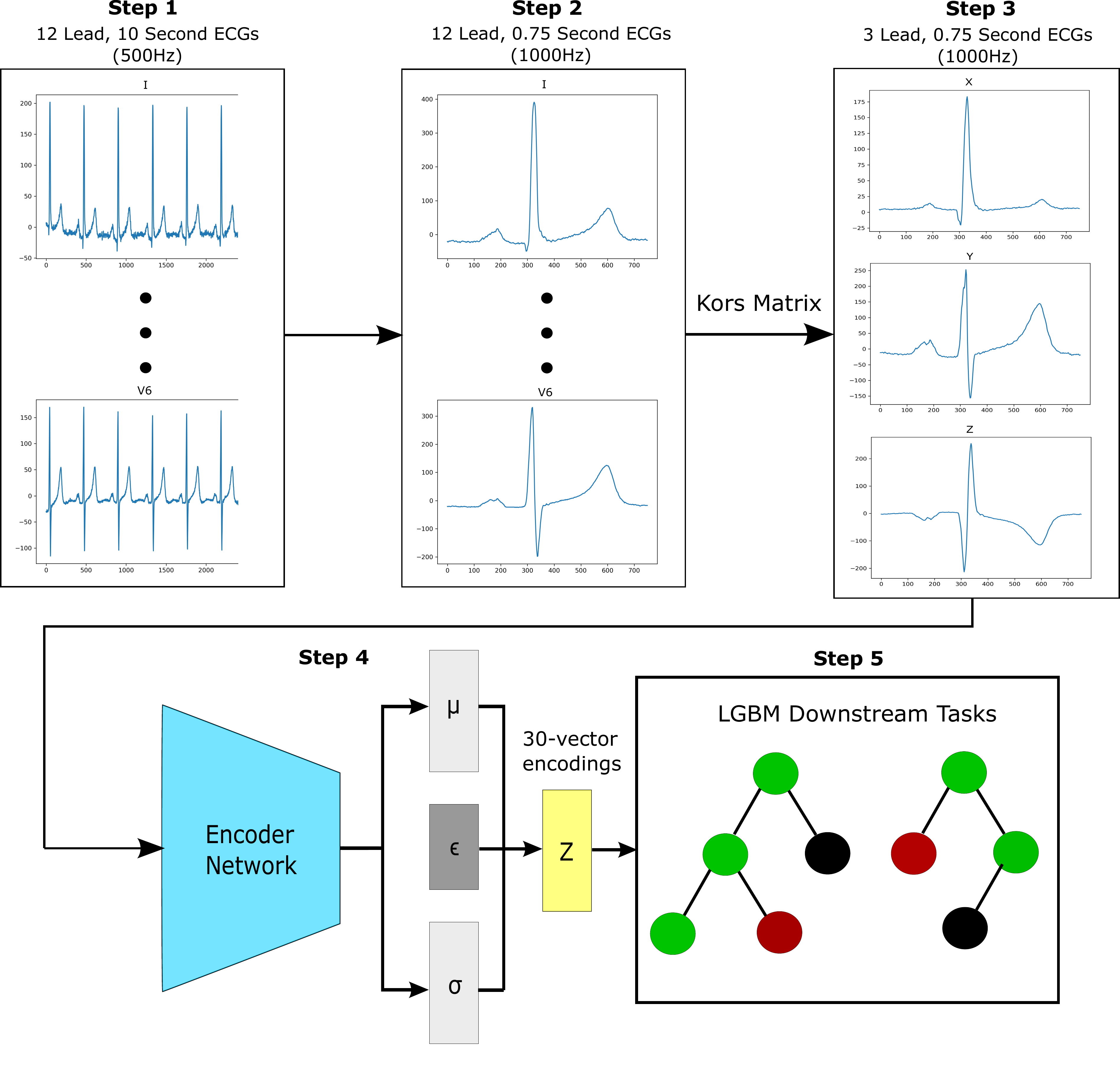}
  
  \caption{Data pipeline for downstream prediction tasks. Converting from 10 s data to 0.75 s to X,Y,Z beats to 30-vector encoding.}
  \label{fig:pipeline}
\end{figure}

To further reduce the data, we applied the Kors’s conversion matrix \cite{kors1990reconstruction} to the eight independent ECG leads (I, II, V1-V6) to obtain orthogonal X (right to left), Y (superior to inferior) and Z (anterior to posterior) lead representations in 3-dimensional space.
This transformation simplifies the data by projecting the 12 ECG leads onto three orthogonal axes. The Kors's conversion matrix can be applied in the backward direction to reobtain the original 12-lead ECG.
Sampled at 1000 Hz, a 120,000-datapoint 10-sec 12-lead ECG is thus reduced to 2250-datapoint 750-ms 3-lead ECG [Fig. \ref{fig:pipeline}].
For stability during model training, we then applied min-max scaling to normalize the amplitude of the data, ensuring that all values fell within a range of -1 to 1.

Our dataset consisted of 1,065,368 ECG signals, which we used to train the encoding models.
For training and testing downstream clinically relevant predictions, we split the data by unique patient medical record numbers into a training set (90\%) and a holdout test set (10\%), ensuring that the models were evaluated on data from unseen individuals.

\textbf{IRB Approval:} This project was performed under the approval as an expedited review by the Institutional Review Board at The University of Kansas Medical Center (STUDY00160252).
The ECGs for this project encompassed all clinically acquired ECGs at The University of Kansas Health System (2008-2022).
ECGs were linked to variables obtained on clinical echocardiographic evaluations, when available, via Healthcare Enterprise Repository for Ontological Narration (HERON) linked by medical record numbers\cite{Murphy2010}\cite{Waitman2011}.

\subsection{Overview of the Models}

We compare 7 models to reduce the dimensionality of ECG data: PCA, AE, Stochastic Autoencoder (SAE), VAE, $\beta$-VAE, cyclical $\beta$-VAE (C$\beta$-VAE), and Annealed $\beta$-VAE (A$\beta$-VAE).
The SAE, C$\beta$-VAE, and A$\beta$-VAE are novel implementations of VAEs, each with unique features tailored to address challenges in ECG data encoding.

The PCA model was trained using the Incremental PCA method from the sci-kit learn library, was configured to produce 30 components to match the 30 latent encodings generated by the VAEs, enabling a fair comparison between different methods.
The AE and its variants (SAE, VAE, $\beta$-VAE, C$\beta$-VAE, A$\beta$-VAE) were trained using the same architecture, with variations in the loss function tailored to each specific model, as discussed in subsequent sections.

\subsection{Overview of VAEs}

All VAE models used in this study share a common architecture, consisting of an encoder and a decoder, both implemented as Convolutional Neural Networks (CNNs).
The encoder network comprises four 2D convolutional layers with filter sizes of 256, 256, 512, and 512, sequentially.
Each layer uses a filter width of 9, a stride of 2, and TanH activation functions.
Batch normalization is applied after each activation to stabilize the training process and improve model generalization.
The final 2D convolutional layer is flattened and passed to two fully connected layers with L2 regularization, 0.01, on the weights and TanH activations with dropout, 0.25, between them.

\begin{figure}  
  \centering
  \includegraphics[width=0.8\linewidth]{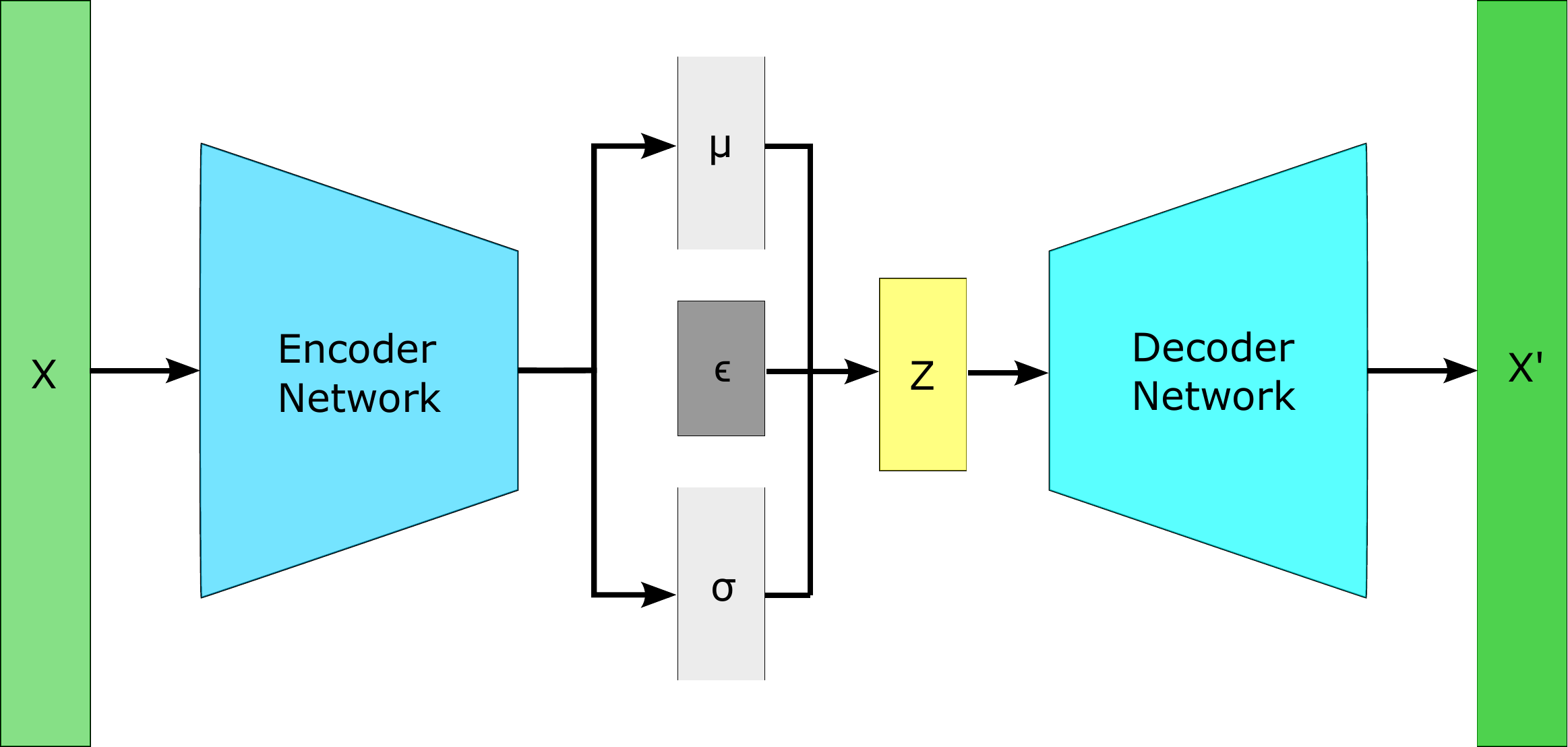}
  
  \caption{Overview of VAE architecture}
  \label{fig:modelarch}
\end{figure}

The final fully-connected layer is sampled using the population mean and log variance of a Gaussian distribution.
This sampling process creates a z-bottleneck where z is defined as:
\begin{equation}
z = \mu + \sigma \odot \epsilon
\end{equation}
Where z is the latent variable encoding, $\mu$ is the population mean, $\sigma$ is the standard deviation of the population distribution, and $\epsilon$ is a small randomly sampled value to allow for back propagation via the reparameterization trick\cite{kingma2013auto}.
The encodings, z, are then passed to the decoder CNN which is made up of 2 fully connected layers with L2 regularization and TanH activations and dropout between.
Followed by four 2D transpose convolutional layers with filters 512, 256, 128, 1 and filter widths of 9 and strides of (1; 1; 1,2; 1,2) with batch normalization between layers.
The final transpose layer recreates the original signal.
The general model architecture can be seen in Fig. \ref{fig:modelarch}.

The use of TanH, batch normalization, and L2 regularization was to zero-base the encodings activations and encodings rather than doing post-processing to normalize the encodings.
This makes any extreme variance from 0 a significant event in the encodings and allows future models to predict based on the variance and distribution of the encodings.
We used Tensorflow for both data preprocessing and the model architecture.
We used Adam for the optimization function.
The model used a learning rate of 0.000001.
It was trained for 50 epochs and had a batch size of 32.
We tested different values for strides (1, 2, 4, mixed), filters (16, 32, 64, 128, 256, 512), max and average pooling, activation functions (ReLU, Leaky ReLU, TanH), epochs (10, 20, 30, 40, 50), learning rates (fixed 1e-5, 1e-6, 1e-7, cyclical, reduce on plateau), batch sizes (8, 16, 32, 64, 128), latent variables (10, 20, 24, 30, 32, 45, 50, 100).
We tried to include variational dropout as well \cite{kingma2015variational}.
This had no effect on the training loss or encodings.
We also tried Lion \cite{chen2023symbolic} for the optimizer which got good results, but was less stable in our environment (Windows 11) than Adam in Tensorflow 2.10.
Iterating was relatively slow with this model.
It took around 2 days to fully train a model for comparison using a GeForce RTX 3090.
The values reported in this paper produced the best result that we tested (not exhaustive).
A larger study with a more comprehensive grid search would be needed to determine the absolute best hyperparameters.

We found that 100-300k ECGs are sufficient to train the encoder.
More training data only slightly improved the fidelity of the reconstructions from the encodings.
Regardless, we trained the models on over 1 million ECGs.

\subsection{Modified ELBO Loss}

The reconstruction process is guided by a Evidence Lower Bound (ELBO) loss function that combines a weighted Mean Squared Error (MSE) between the original and reconstructed signals, a Kullback–Leibler (KL) divergence term to regularize the latent space \cite{kullback1951information}, and a beta term to control the balance between reconstruction quality and feature disentanglement.
Each 250-ms section of the signal (P wave, QRS complex, and T wave) has unique weights that are selected to diminish the effects of differences in their amplitudes.
This allows the model to focus evenly across all the waves and accurately reconstruct all of them without giving priority to the higher amplitude QRS complex.

The loss function $L_\beta$ is given below
\begin{equation}
L_\beta = L_E + \beta KL
\end{equation}
\begin{equation}
L_E = \theta_{\text{QRS}} \cdot L_{\text{QRS}} + \theta_{\text{T}} \cdot L_{\text{T}} + \theta_{\text{P}} \cdot L_{\text{P}}
\end{equation}
\begin{equation}
KL = -0.5 \left(1 + \ln(\sigma^2_z) - (\mu_z)^2 - e^{\ln(\sigma^2_z)} \right)
\end{equation}

Where $L_\beta$ is the total loss of the model by which the model's gradient is updated.
$L_E$ is the weighted MSE of the model where $L_{\text{QRS}}$ is the MSE between the original signal, $x$ and its reconstructed output $x'$.
$L_{\text{T}}$ and $L_{\text{P}}$ are the same for the T and P waves respectively.
$\theta_{\text{QRS}}$, $\theta_{\text{T}}$, and $\theta_{\text{P}}$ are scalar weights to augment the importance of each segment of the ECG.

We initially trained this VAE just using MSE as the reconstruction loss function and we found that the model focused on the high-amplitude QRS complex and was unable to reconstruct the P and T waves.
The QRS complex generally has 2-10x greater amplitude compared to P and T waves, with the P wave having the smallest amplitude.
We added weights to each segment of the signal to make the model give more importance to the P and T waves.
We tried different weights and found that as long as the QRS complex weight is $\leq$ half of the P wave weight the model can successfully reconstruct all 3 segments of the signal.
The weights used in this paper are $\theta_{\text{P}}=20.0$, $\theta_{\text{QRS}}=10.0$ and $\theta_{\text{T}}=15.0$.
We included a weight for the QRS complex as it made the model more aggressive during the AE training epochs and helped with convergence.

The AE is a normal AE with a bottleneck layer to extract the signal representations for downstream prediction.
It had the same encoder and decoder architecture as above just without the $z$, $\mu$, and $\sigma$ components.
It was trained solely on the weighted MSE reconstruction loss.

$KL$ is the Kullback–Leibler loss for Gaussian distributions.
The KL loss measures the distance between the approximate posterior distribution q($z|x$) and the prior distribution p($z$) over the latent variable $z$ 
\cite{kullback1951information}.
It encourages the latent representations to be similar to the prior distribution, promoting smoothness and continuity in the latent space

$\beta$ is the scaling factor for the $KL$ loss.
The higher the $\beta$ value, the greater the control of the generated reconstructions becomes, but the worse the reconstruction quality gets. 
A higher $\beta$ encourages disentanglement and feature distinction.
If the encodings are disentangled, then each encoding affects just one aspect of the signal.
Disentanglement between the encodings is useful for generating realistic synthetic data using the VAE by synthetic oversampling techniques.

For the $\beta$-VAE, the value of $\beta$ was set to 3.

\subsection{Overview of Novel Variants}

For the C$\beta$-VAE, we implemented a cyclical annealing schedule where the values of $\beta$ range from 0 to 5, changing each epoch during training.
This cycling means that the model alternates between three different loss functions: AE ($\beta$ = 0), VAE ($\beta$ = 1), and $\beta$-VAE ($\beta$\textgreater1).
At $\beta$=0, the model is essentially just an AE where the loss function only includes the reconstruction error, $L_E$.
At $\beta$=1, the model is a regular VAE with a KL loss and error term\cite{kingma2013auto}.
At $\beta$\textgreater1, there is an additional term added to increase the model’s focus on disentanglement by enhancing the effect of KL on the total loss\cite{higgins2017betavae}.
By cycling from 0 to 5 we make the model go through periods of focusing purely on reconstruction and periods where the model focuses more on understanding the abstract interaction between the data points.
The original paper which introduced cyclical annealing KL loss \cite{fu2019cyclical} had the $\beta$ term cycle between 0 and 1.
They also had $\beta$ hard reset back to 0 from 1.
Instead, we propose to have the $\beta$ term go between 0 and 5 without a hard reset.
$\beta$ goes from 0 to 5 in 10 epochs and 5 to 0 in 10 epochs with a complete cycle  every 20 epochs.
This cyclical annealing approach accelerates convergence, enabling the model to produce usable reconstructions within just 10 epochs.
Further experimentation is needed to determine whether this cyclical approach is superior to the hard reset method.

The A$\beta$-VAE, on the other hand, utilizes a reverse annealing process.
In this variant, the $\beta$ value starts at 10 and is gradually reduced to 0 each epoch over the course of 50 epochs.
This process allows the model to initially focus on disentanglement and the structure of the latent space before shifting its emphasis toward reconstruction.
By beginning with a high $\beta$ value, the A$\beta$-VAE emphasizes the regularization of the latent space, which can lead to more meaningful and distinct features in the encoded representations.
This produced a model which was exceptionally good at reconstruction fidelity.

For the SAE, the value of $\beta$ was set to 0, which omits the KL term altogether.
This essentially creates an AE with a stochastic latent distribution, z, which is only trained using the reconstruction loss.
So, instead of learning a Gaussian distribution from the $KL$ loss, it learns the distribution which minimizes the reconstruction loss.
The purpose of the VAE architecture was to create a Gaussian distribution for data generation purposes.
The SAE is a counter-intuitive departure from that to create an AE with generative capabilities which only focuses on the reconstruction of the ECG signal.
The SAE allows for a more flexible and data-driven approach to encoding which we can see improves the performance of downstream tasks while also maintaining the advantages of VAEs in data synthetic oversampling generation methods.

\begin{table*} [h]
\caption{MAE, MSE, and DTW for Different Models for Representative Beat X, Y, Z-Lead ECG Reconstructions (N=1,065,368)}
\centering
\scriptsize
\begin{tabular}{lcccccc}
\hline
Model & 1st 250 ms (p wave) & 2nd 250 ms (QRS) & 3rd 250 ms (T wave) & Full signal MAE & Full signal MSE & Full signal DTW \\
& MAE ($\mu V$) Avg $\pm$ SD & MAE ($\mu V$) Avg $\pm$ SD & MAE ($\mu V$) Avg $\pm$ SD & MAE ($\mu V$) Avg $\pm$ SD & ($\mu V^2$) Avg $\pm$ SD & Avg $\pm$ SD \\
\hline
PCA & 19.1$\pm$6.5 & 29.5$\pm$7.3 & 22.7$\pm$7.7 & 24.0$\pm$5.0 & 1842.9$\pm$840.0 & 667.9$\pm$218.7 \\
AE & \textbf{11.2$\pm$3.0} & 23.2$\pm$5.5 & \textbf{12.8$\pm$3.7} & 15.8$\pm$3.1 & 739.8$\pm$316.1 & 313.7$\pm$94.0 \\
SAE & 11.4$\pm$2.7 & 31.7$\pm$8.1 & 12.7$\pm$3.4 & 18.7$\pm$3.7 & 1131.3$\pm$514.8 & 387.3$\pm$131.8 \\
VAE & 11.9$\pm$2.7 & 28.9$\pm$6.9 & 12.8$\pm$3.3 & 17.9$\pm$3.5 & 996.2$\pm$433.2 & 361.2$\pm$115.5 \\
$\beta$-VAE & 11.5$\pm$3.0 & 23.6$\pm$5.6 & 13.1$\pm$3.7 & 16.2$\pm$3.2 & 755.6$\pm$325.6 & 317.5$\pm$94.8 \\
A$\beta$-VAE & \textbf{11.2$\pm$3.0} & \textbf{22.6$\pm$5.3} & \textbf{12.8$\pm$3.8} & \textbf{15.7$\pm$3.2} & \textbf{701.6$\pm$304.8} & \textbf{308.1$\pm$92.1} \\
c$\beta$-VAE & 12.0$\pm$2.6 & 31.9$\pm$6.6 & 14.0$\pm$3.8 & 19.3$\pm$3.4 & 1202.7$\pm$491.3 & 400.6$\pm$130.3 \\
\hline
\end{tabular}
\label{tab:MAE_MSE_DTW}
\end{table*}

\begin{table*}[h] 
\caption{Comparison of MAE and DTW for X, Y, and Z Signals Across Models (N=97,464)}
\centering
\begin{tabular}{lcccccc}
\hline
Model & X Signal MAE ($\mu V$) & Y Signal MAE ($\mu V$) & Z Signal MAE ($\mu V$) & X Signal DTW & Y Signal DTW & Z Signal DTW \\
\hline
PCA & 25.8$\pm$21.4 & 25.6$\pm$20.7 & 23.5$\pm$21.0 & 751.9$\pm$856.4 & 715.8$\pm$778.0 & 647.2$\pm$766.0 \\
AE & 16.3$\pm$13.4 & 16.9$\pm$13.1 & 16.5$\pm$13.2 & 338.0$\pm$412.4 & 337.0$\pm$384.9 & 332.6$\pm$390.8 \\
SAE & 19.7$\pm$12.8 & 21.1$\pm$12.7 & 17.1$\pm$11.9 & 429.5$\pm$404.8 & 455.9$\pm$406.5 & 340.0$\pm$362.3 \\
VAE & 18.9$\pm$14.0 & 20.5$\pm$14.4 & 16.4$\pm$11.8 & 400.4$\pm$413.1 & 419.5$\pm$400.1 & 327.9$\pm$349.7 \\
$\beta$-VAE & 16.9$\pm$14.0 & 17.8$\pm$14.0 & \textbf{15.9$\pm$12.6} & 346.5$\pm$413.8 & 350.4$\pm$390.2 & 321.4$\pm$376.6 \\
A$\beta$-VAE & \textbf{16.1$\pm$13.5} & \textbf{16.9$\pm$12.9} & 16.0$\pm$13.1 & \textbf{334.3$\pm$411.2} & \textbf{335.8$\pm$384.4} & \textbf{321.7$\pm$381.1} \\
c$\beta$-VAE & 20.6$\pm$14.9 & 22.0$\pm$15.2 & 17.8$\pm$11.5 & 438.2$\pm$425.6 & 462.3$\pm$422.5 & 365.5$\pm$358.2 \\
\hline
\end{tabular}
\label{tab:MAE_DTW_Signals}
\end{table*}

\subsection{Comparison of Models for Downstream Prediction}

We used prediction of reduced LVEF ($\leq$35\%) to compare downstream clinically relevant predictions.
The baseline comparator CNN on the full 10-sec 12-lead ECG data we used for this paper is our implementation of the CNN model architecture from Mayo Clinic\cite{attia2019screening}.
This is a state-of-the-art CNN architecture which has been used to model a variety of targets (sex, age, LVEF, etc\cite{attia2019screening}\cite{attia2019age}).
The performance of our CNN model for reduced LVEF in our test set has AUROC 0.909 which is in the range published by other centers \cite{cho2021artificial}.

We compare the performance of our baseline CNN model trained on raw ECG signal data with Light Gradient Boosted Machine (LGBM) models \cite{10.5555/3294996.3295074} trained on the encodings from our AE models.
For the LGBM models, We only altered 8 parameters from their default values: Max Depth (15), Colsample Bytree(.9), Extra Trees (True), Top K (100), Learning Rate (0.1), Num Estimators (1,000,000 with early stopping), Reg Alpha/Lambda (0.95).

\section{Experiments and Results}

\subsection{Signal Reconstruction}

\begin{figure*}[h]  
  \centering
  \includegraphics[width=1.\linewidth]{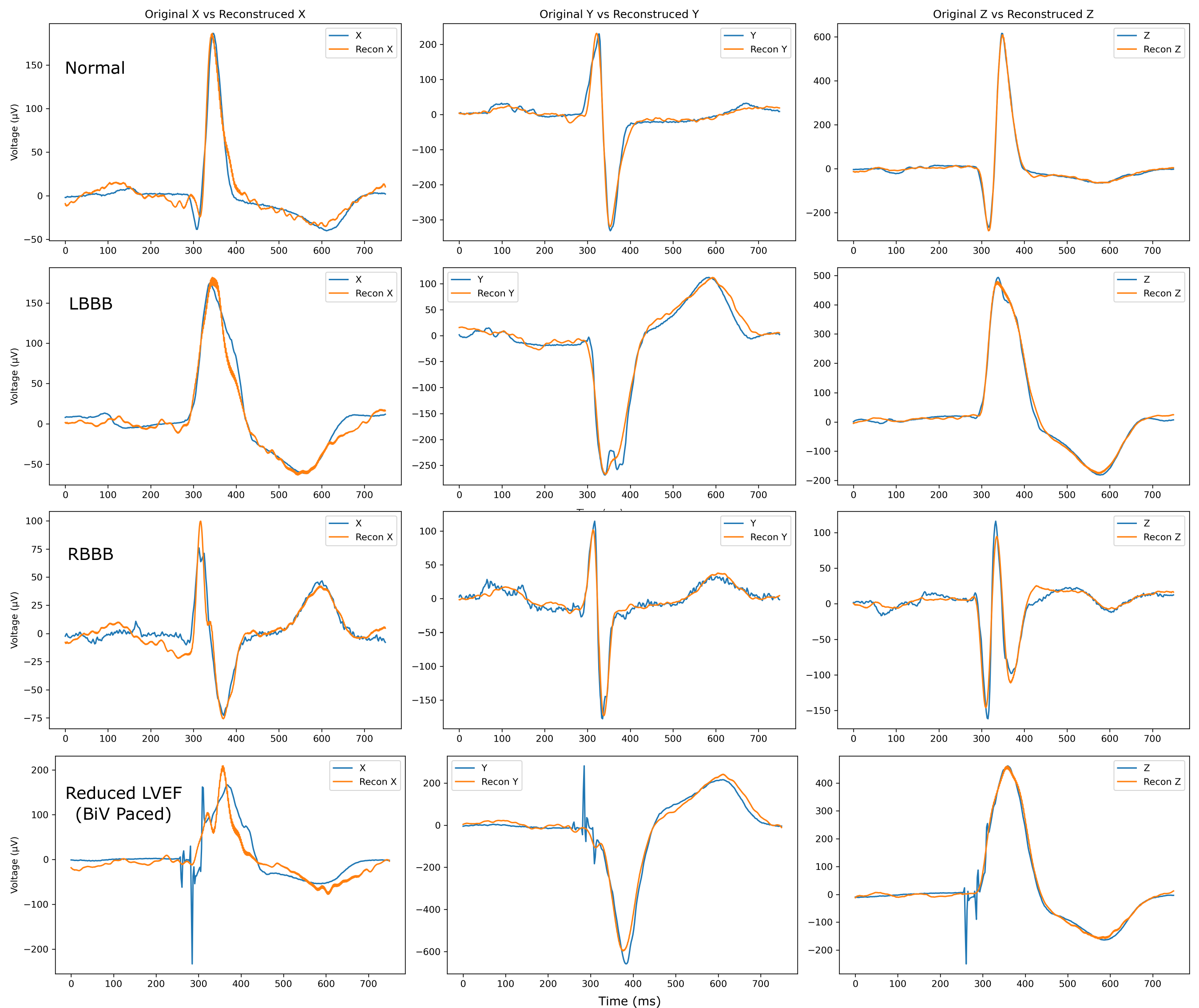}
  
  \caption{4 examples (Normal, LBBB, RBBB, and reduced LVEF) of signal reconstruction (orange) with original signal (blue) overlaid from the A$\beta$-VAE model. The model can reconstruct with a variety of noises and generally smooths out the artifacts.}
  \label{fig:sigrecon}
\end{figure*}

We compare each of the 7 models for their ability to reconstruct each subsection of the ECG signal in Table \ref{tab:MAE_MSE_DTW} using the Mean Absolute Error (MAE), Mean Squared Error (MSE), and Dynamic Time Warping (DTW) score.
DTW measures the similarity between two temporal sequences by aligning them non-linearly to account for variations in timing or speed, making it ideal for comparing sequences with temporal distortions.
PCA reconstruction was measured via the inverse transform function.

All of the AE/VAEs got good results with the Annealing $\beta$-VAE achieving the best results across all metrics, closely followed by pure AE.
Annealing $\beta$-VAE had a mean average error of 15.7 $\mu$V which is around the noise level of these signals.
The average QRS amplitude for comparison was 942.8$\pm$434.2 $\mu$V.
This shows that the signals can be accurately reconstructed via all of the VAEs but the process of annealing the $\beta$ value improves the reconstruction fidelity.
The signal fidelity was even better than a pure AE which is designed only to recreate the original signal.
PCA was significantly worse than any of the AEs with its DTW being more than twice that of A$\beta$-VAE.

Examples of the original signal and the reconstruction from the A$\beta$-VAE model overlaid across varied ECG types are shown in Fig. \ref{fig:sigrecon}. 
The model operates by reconstructing the X, Y, and Z leads concurrently, using the same set of 30 latent variables.
The ability of the model to use only 30 data points to effectively reconstruct all 3 leads demonstrates the efficiency of the latent representation.
We observed differences in reconstruction error among the leads, with the Z lead often showing distinctive higher quality reconstructions compared to the X and Y leads.
The models on average reconstruct the Z lead 11.2\% better than the X lead and 12.5\% better than the Y lead.
The differences can be seen in Table \ref{tab:MAE_DTW_Signals}.

A possible explanation for the better reconstruction of Z lead from the VAE encodings may be related to the heart-torso anatomic relationship and the way ECG is recorded. 
The Z lead may exhibit unique amplitude and morphological features due to its relation to the anatomic cardiac positioning.
It may be impacted by the higher weightage of the unipolar chest leads V1-V6 as compared to bipolar limb leads I and II in Kors conversion matrix \cite{kors1990reconstruction}.
The chest leads are notable to have less artifact compared to limb leads.

The QRS complex has a larger error compared to P and T waves.
This is naturally due to the larger QRS amplitude and the higher weights for P and T wave in the reconstruction loss function.

\subsection{Prediction of QRS Measurements}

\begin{table}[h]
\caption{Prediction in Test Set of ECG Measurements With LGBM Using Representative Beat X, Y, Z-Lead ECG Encoded Variables (N=97,464)}
\centering
\scriptsize
\setlength{\tabcolsep}{3pt} 
\begin{tabular}{lcccccc}
\hline
Model & \multicolumn{2}{c}{QRS Duration (ms)} & \multicolumn{2}{c}{Amplitude$_{\text{QRS-3D}}$ ($\mu V$)} & \multicolumn{2}{c}{VTI$_{\text{QRS-3D}}$ ($\mu Vs$)} \\
\cline{2-3} \cline{4-5} \cline{6-7}
& MAE $\pm$ SD & $R^2$ & MAE $\pm$ SD & $R^2$ & MAE $\pm$ SD & $R^2$ \\
\hline
PCA & \textbf{8.1$\pm$12.9} & \textbf{0.727} & \textbf{108.3$\pm$173.3} & \textbf{0.838} & \textbf{3.16$\pm$5.32} & \textbf{0.923} \\
AE & 8.7$\pm$13.8 & 0.687 & 108.8$\pm$175.3 & 0.835 & 3.68$\pm$5.96 & 0.903 \\
SAE & 8.3$\pm$13.2 & 0.712 & 109.1$\pm$176.6 & 0.835 & 3.42$\pm$5.64 & 0.914 \\
VAE & 8.4$\pm$13.4 & 0.705 & 109.7$\pm$177.6 & 0.833 & 3.46$\pm$5.72 & 0.911 \\
$\beta$-VAE & 8.5$\pm$13.5 & 0.697 & 109.6$\pm$177.4 & 0.833 & 3.54$\pm$5.83 & 0.908 \\
A$\beta$-VAE & 8.3$\pm$13.3 & 0.709 & 107.7$\pm$175.3 & 0.837 & 3.44$\pm$5.68 & 0.912 \\
c$\beta$-VAE & 8.3$\pm$13.3 & 0.708 & 110.2$\pm$177.9 & 0.832 & 3.44$\pm$5.69 & 0.912 \\
\hline
\end{tabular}
\label{tab:ECG_Predictions}
\end{table}

To validate that the encodings from 7 models, by themselves, can predict direct automated measurements off the ECG signal, we trained separate LGBMs for 3 different QRS measurements from only the 30 encodings from each model as the input. 
The encodings can predict the QRS duration, the 3D QRS amplitude, and the scalar 3D QRS voltage-time integral VTI$_{\text{QRS-3D}}$.
The results for this can be seen in Table \ref{tab:ECG_Predictions}.
For reference, the average measured QRS duration was 94.8$\pm$29.2 ms, amplitude$_{\text{QRS-3D}}$ was 942.8$\pm$434.2 $\mu$V and the VTI$_{\text{QRS-3D}}$ was 32.12$\pm$17.43 $\mu$Vs, and these were predicted from the encodings with an error  in the range of approximately 10\%.
PCA performed the best in this task with R\(^2\) to predict VTI$_{\text{QRS-3D}}$  0.923. This isn't surprising as PCA is extracting separable features from the signal data which likely closely relate to the original features.
The best performing VAE was the SAE with R\(^2\) to predict VTI$_{\text{QRS-3D}}$  0.914.

\subsection{Prediction of Bundle Branch Blocks}

The right and left bundle branches are the main conduction branches for ventricular activation.
Conduction delay or block in either of these results in Right Bundle Branch Block (RBBB) or Left Bundle Branch Block (LBBB) respectively.
RBBB and LBBB lead to increased QRS duration, both with distinctive QRS morphological alterations.
We used the American Heart Association/American College of Cardiology Foundation/Heart Rhythm Society criteria for diagnosis of RBBB and LBBB to develop custom code to identify these conduction abnormalities \cite{Surawicz2009}.

\begin{table}[h]
\caption{Prediction in Test Set of RBBB and LBBB With LGBM Using Representative Beat X, Y, Z-Lead ECG Encoded Variables (N=97,464)}
\centering
\scriptsize
\begin{tabular}{lcccc}
\hline
Model & \multicolumn{2}{c}{RBBB (8.06\% Prevalence)} & \multicolumn{2}{c}{LBBB (3.99\% Prevalence)} \\
\cline{2-3} \cline{4-5}
& AUROC & Sensitivity/Recall & AUROC & Sensitivity/Recall \\
\hline
PCA & 0.9435 & 0.894 & 0.9637 & 0.939 \\
AE & 0.9390 & 0.881 & 0.9618 & 0.938 \\
SAE & 0.9504 & 0.906 & \textbf{0.9701} & \textbf{0.948} \\
VAE & 0.9507 & 0.904 & 0.9688 & 0.950 \\
$\beta$-VAE & 0.9473 & 0.895 & 0.9689 & 0.949 \\
A$\beta$-VAE & 0.9499 & 0.903 & 0.9686 & 0.947 \\
c$\beta$-VAE & \textbf{0.9516} & \textbf{0.908} & 0.9697 & 0.949 \\
\hline
\end{tabular}
\label{tab:RBBB_LBBB}
\end{table}

We trained separate LGBM models using the 30 encoded variables alone from each of the 7 ECG reducing models to predict RBBB and LBBB. 
Table \ref{tab:RBBB_LBBB} shows the LGBM models' predictive performance to  classify RBBB and LBBB.
The RBBB and LBBB labels are calculated with a complex combination of QRS characteristics of the ECG signal.
PCA and pure AE were inferior to all VAEs for prediction of these labels with complex calculated features of the data.
Despite the imbalanced distribution of the RBBB/LBBB labels, all VAE encodings performed well.
The AUROCs for either label with LGBM models on latent encodings from all VAEs were approximately 0.950 or higher, with sensitivity/recall (at specificity fixed at 0.900) of approximately 0.900 or higher.
The encodings from c$\beta$-VAE were the best for predicting RBBB with an AUROC of 0.9516.
The SAE encodings had the best model for predicting LBBB with an AUROC of 0.9701.

\subsection{Autoencoder Latent Variables as Surrogate for ECG Signal}

The high-fidelity of signal reconstruction, and the high predictive ability of the VAE encodings for identifying ECG features including QRS measurement and morphological characteristics shows that the encodings are able to be a good surrogate for the raw signals.
This is an improvement of representation over summary statistics as it can not only represent inherent ECG features but also capture subtle morphological variations in the ECG signal to allow reconstruction of the original signal.
In addition to the dimensionality reduction allowing utilization of these encodings for simpler and faster learning models, e.g., tree-based or simple ANNs, access to such subtle ECG signal characteristics is crucial in making clinically-relevant predictions. 

With the use of tree-based models, such as LGBM, we can also retrieve the feature importance of the model to see which specific encodings have the most use during training.
Though the individual encodings used for each specific model varies, we found that the models tend to use all 30 encodings in roughly the same number of trees.
This suggests that each encoding holds some disentangled intrinsic value that the models use to generate predictions without relying heavily on any individual encoding.

\subsection{Downstream Clinically-relevant Prediction}
\begin{table}[h]
\caption{Prediction in Test Set of Reduced LVEF With LGBM Using Representative Beat X, Y, Z-Lead ECG Encoded Variables in Test Set (n=30,554)}
\centering
\scriptsize
\begin{tabular}{lcccc}
\hline
Model & \multicolumn{2}{c}{Reduced LVEF (14.09\% Prevalence)} & \multicolumn{2}{c}{LVEF, \%} \\
\cline{2-3} \cline{4-5}
& AUROC & Sensitivity/Recall & MAE $\pm$ SD & $R^2$ \\
\hline
PCA & 0.799 & 0.616 & \textbf{8.86$\pm$12.15} & \textbf{0.247} \\
AE & 0.810 & 0.656 & 9.05$\pm$12.42 & 0.213 \\
\textbf{SAE} & \textbf{0.820} & \textbf{0.665} & 8.96$\pm$12.28 & 0.231 \\
VAE & 0.819 & 0.676 & 8.95$\pm$12.26 & 0.233 \\
$\beta$-VAE & 0.812 & 0.663 & 9.05$\pm$12.39 & 0.217 \\
A$\beta$-VAE & 0.818 & 0.666 & 8.96$\pm$12.29 & 0.229 \\
c$\beta$-VAE & 0.820 & 0.675 & 8.97$\pm$12.28 & 0.231 \\
ECG statistics& 0.761& 0.554 & 9.55$\pm$12.92 & 0.148 \\
\hline
\end{tabular}
\label{tab:LVEF_Prediction}
\end{table}

To make comparisons, we used reduced LVEF as the clinically-relevant downstream prediction.
LVEF is the percent of blood that is pumped by the left ventricle with each contraction.
We had 303,265 ECGs on 105,370 patients paired with a unique echocardiogram within 45 days showing reduced LVEF ($\leq$35\%) in 14.1\%.

We trained an LGBM model on each of the model encodings on both linear prediction of LVEF \% values and bivariate classifier for LVEF $\leq$35\%, Table \ref{tab:LVEF_Prediction}.
The SAE encodings achieved the best performance on the binary classification task, with an AUROC of 0.8203, closely followed by the C$\beta$-VAE encodings with an AUROC of 0.8201.
In comparison, models trained using only basic ECG summary statistics (e.g., heart rate, PR interval, QRS duration, QRS axis, corrected QT interval) achieved an AUROC of 0.7605 and PCA got AUROC 0.799, demonstrating the superior predictive capability of the VAE encodings.

To obtain a better model for reduced LVEF detection, we combined the SAE encodings with simply extracted ECG features—including heart rate, PR interval, QRS duration, QRS axis, corrected QT interval, T peak-to-T end duration, amplitudes of various waves (QRS, R, S, T), and voltage-time integrals (QRS and QRS-T) from different leads—in an LGBM model.
This combined model achieved an AUROC of 0.901 in the independent holdout test set (n=30,554).
As a comparison, the best standard convolutional neural network model on the full raw 12-lead ECG signal data (analogous to the state-of-the-art published model \cite{attia2019screening}) achieved a test set AUROC of 0.909.

\begin{table}[h]
\caption{Performance of Different Machine Learning Models in Predicting Reduced LVEF (LVEF$\leq$35\%) from ECG Data (Holdout Test Set: n=15,987)}
\centering
\scriptsize
\begin{tabular}{lccc}
\hline
Model & Training Sample Size & AUROC & Sensitivity/Recall \\
\hline
CNN & 100\% (n=143,644) & \textbf{0.909} & \textbf{0.742} \\
 & ~9.5\% (n=13,568) & 0.630 & 0.177 \\
ResNet & 100\% (n=143,644) & \textbf{0.892} & \textbf{0.672} \\
 & 10\% (n=14,364) & 0.855 & 0.586 \\
 & 1\% (n=1,436) & 0.811 & 0.462 \\
 & 0.1\% (n=143) & 0.705 & 0.281 \\
LGBM & 100\% (n=143,644) & \textbf{0.901} & \textbf{0.702} \\
 & 10\% (n=14,364) & 0.870 & 0.610 \\
 & 1\% (n=1,436) & 0.846 & 0.525 \\
 & 0.1\% (n=143) & 0.761 & 0.361 \\
\hline
\end{tabular}
\label{tab:reduced_size}
\end{table}

As a proof of concept to evaluate the robustness of  LGBM models using SAE encodings with lesser amounts of training data, we conducted experiments with reduced training set sizes \cite{HarveyOverfitting}.
Table \ref{tab:reduced_size} summarizes the performance of different models in predicting reduced LVEF across different training sample sizes. The table compares the CNN model on full raw ECG signal with a fully-connected (non-convolutional) residual network (ResNet) and LGBM models on SAE encodings and the simple ECG measurements/features. 

With the full training set, the LGBM model achieved an AUROC of 0.901 approaching the CNN's performance (0.909).
When the training data was reduced to 10\%, however, the CNN's performance dropped significantly (AUROC of 0.630), indicating overfitting due to insufficient training data.
In contrast, the LGBM model maintained robust performance (AUROC of 0.870).
Even with an extremely small training dataset of only 143 samples, the LGBM model achieved an AUROC of 0.761. This demonstrates the effectiveness of VAE encodings in dimensionality reduction and underscores the practical utility of the ECG encoded representations for machine learning when large-scale labeled data is not available.

These results highlight that VAE encodings allow traditional machine learning models using substantially less training data and computational resources to perform comparably to DL models trained on full ECG signals.
This approach would facilitate the development of ECG diagnostic tools for minority populations, rare health conditions, and invasive modalities where large training datasets are not unavailable.

\subsection{VAE Performance}

The overall best performing VAE variant for downstream prediction tasks was the SAE. This result is unexpected because, unlike VAEs, which use the KL divergence to regularize the latent space and enforce a structured representation towards a prior distribution, the SAE relies solely on reconstruction loss.
In the SAE, without the KL divergence term, the model cannot enforce any similarity amongst the latent variables and regularize the latent space distribution towards a prior (such as a standard normal distribution).
Consequently, the z mean and z variance parameters are optimized solely based on the reconstruction loss, without any regularization enforcing a specific latent distribution.
This approach typically does not promote disentanglement or feature separation in the latent space, often leading to poor generalization.
However, the SAE outperformed the VAE model, despite the VAE being traditionally favored for its ability to balance reconstruction quality with a structured latent space.

These findings suggest that the regularization of the latent space by KL loss is not as important for ECG encoding.
Given that ECG signals are inherently highly variable, it is likely that this natural variability aligns better with the unregularized latent space of the SAE.
The absence of the KL term might, in this case, be beneficial, as it allows the model to focus entirely on reconstructing the high variability of the ECG signals without enforcing a potentially limiting structure on the latent space.
This suggests that for highly variable data like ECGs, the additional regularization imposed by the KL divergence might not always be necessary and could, in fact, hinder downstream predictions.
The SAE’s success underscores the trade-off between reconstruction fidelity and latent space regularization, showing that precise reconstruction can sometimes be more valuable than a structured latent space, particularly for data with significant inherent variability.

On the other hand, the AE was decidedly outperformed by all VAEs including SAE for all prediction tasks. This demonstrates that the generalizability from introducing the stochastic sampling from a distribution to obtain the latent variables in the VAE architecture is still important for downstream predictions from ECG. 

\subsection{Discussion}

Our study presents significant advancements in applying autoencoder techniques, particularly our novel VAE variants, to ECG data analysis in clinical diagnostics.
By effectively reducing high-dimensional ECG signals to a compact set of latent variables without requiring extensive datasets, we address a critical challenge in utilizing ECG data in machine learning models.
Our novel VAE variants, particularly the A$\beta$-VAE and SAE, demonstrate superior performance in preserving essential morphological features of the ECG by capturing non-linear relationships within the data, which is crucial for accurate clinical interpretations. Compared to traditional methods like PCA, our models better retain clinically relevant information, as evidenced by the improved prediction of complex ECG features such as bundle branch blocks and reduced LVEF. This enhanced capability leads to improved diagnostic accuracy, which is particularly impactful for conditions like heart failure where early detection can significantly alter patient outcomes.

By accommodating the inherent variability of ECG signals among individuals—through the SAE's focus on reconstruction quality over latent space regularization—our approach allows for more personalized and accurate assessments, aligning with the move towards personalized medicine.
Additionally, enabling robust predictive models in environments with limited datasets contributes to more equitable healthcare delivery by providing advanced diagnostic capabilities across under-represented and diverse populations.
This extends the utility of VAEs in ECG analysis beyond interpretability to practical predictive performance, offering a pathway to integrate these models into existing clinical workflows even with smaller datasets.

\section{Conclusion and Future work}

This study demonstrates that ECG data, despite its inherent complexity and variability, can be effectively reduced using PCA and VAEs for a wide range of downstream prediction tasks. Our approach shows that 120,000 data points in a full ECG can be reduced to 30 with minimal information loss.
While PCA remains a strong contender for basic feature extraction, its limitations become apparent in more complex prediction tasks where access to the full range of ECG signal details is crucial.
In these scenarios, the novel VAE variants, particularly the SAE and C$\beta$-VAE, demonstrate high-fidelity signal reconstructions and accurate downstream predictions.
The SAE encodings stood out, excelling in all prediction tasks while also providing good signal reconstruction.
This finding challenges the conventional wisdom that regularizing the latent space is always beneficial, suggesting instead that a focus on reconstruction quality can yield equally, if not more, valuable results.
Additionally, VAEs can be used for synthetic signal generation. 

The deficiency of our method is that the encodings lack beat-to-beat information as they are created from one representative averaged heartbeat.
The 10-sec data is crucial for cardiac rhythm and arrhythmic detection and additionally captures information on autonomic nervous function and susceptibility to arrhythmogenesis. 
The next step for complexity reduction research for ECG would be to encode the full 10-sec signal.

By continuing to refine these encoding techniques, we aim to create more robust diagnostic tools that can be applied to minority populations and rare health conditions, ultimately enhancing the accessibility and effectiveness of ECG-based diagnostics. In future work, we will explore clinical applications of these methods in such prediction tasks that currently lack large-scale training datasets. 



\bibliographystyle{IEEEtran}
\bibliography{main.bib}

\end{document}